# Adaptive Local Kernels Formulation of Mutual Information with Application to Active Post-Seismic Building Damage Inference

Mohamadreza Sheibani[1] and Ge Ou[*1]

[1] Department of Civil and Environmental Engineering, University of Utah, Salt Lake City, UT 84112

**Abstract**

The abundance of training data is not guaranteed in various supervised learning applications. One of these situations is the post-earthquake regional damage assessment of buildings. Querying the damage label of each building requires a thorough inspection by experts, and thus, is an expensive task. A practical approach is to sample the most informative buildings in a sequential learning scheme. Active learning methods recommend the most informative cases that are able to maximally reduce the generalization error. The information theoretic measure of mutual information (MI), which maximizes the expected information gain over the input domain, can be used for informative sampling of a dataset in a pool-based scenario. However, the computational complexity of the standard MI algorithm prevents the utilization of this method on large datasets. A local kernels strategy was proposed to reduce the computational costs, but the adaptability of the kernels to the observed labels was not considered in the original formulation of this strategy. In this article, an adaptive local kernels methodology is developed that allows for the conformability of the kernels to the observed output data while enhancing the computational complexity of the standard MI algorithm. The proposed algorithm is developed to work on a Gaussian process regression (GPR) framework, where the kernel hyperparameters are updated after each label query using the maximum likelihood estimation. In the sequential learning procedure, the updated hyperparameters can be used in the MI kernel matrices to improve the sample suggestion performance. The advantages of the proposed method are demonstrated on a simulation of the 2018 Anchorage, AK, earthquake. It is shown that while the proposed algorithm enables GPR to reach acceptable performance with fewer training data, the computational demands remain lower than the standard local kernels strategy.

## 1. Introduction

The normal functionality of the built environment (such as buildings, roads, bridges, power, and communication networks) supports a community's fundamental needs. However, this functionality may be interrupted by major natural and man-made hazards [1]. Especially in a regional hazard, making informed decisions to guide the emergency response or community recovery is challenging. Accurate and promptly gathered knowledge about the distribution of damage in the built environment, including location and severity, facilitates optimal solutions in the evacuation, shelter designation, and financial aid estimations [2].

---

[*] Corresponding Author: ge.ou@utah.edu

## 1.1 Current Practice in Damage Estimation

In the natural hazards research community, state-of-the-art approaches to obtaining the infrastructure's damage condition after regional hazards fit into two categories. The first is through developing physical-law-based functions to map the hazard intensity and infrastructural characteristics to post-hazard performance. These functions can be fragility models describing the probability of the structural responses exceeding a threshold, given the hazard's intensity [3 -6]. Fragility functions can be modeled in two ways, analytically and empirically [7]. Empirical methods rely on the data obtained from historical earthquake events and require expert judgment to classify the observed data for different building types and damage states. The analytical methods, on the other hand, calculate the responses for different groups of engineered buildings and fit fragility curves to the generated data in order to classify the damage state of a building due to seismic demands. However, both analytical and empirical methods require expert judgment and recalibration using historical data in order to be adapted to a specific event.

The other commonly-used approach is deterministic simulations, including finite-element and other simplified models. Simulation-based approaches aim to accurately replicate an individual structure's response time history during an event. In the past decade, with increasing computational capacity and the popularity of the geo-coded information in computer programs (GIS system), single-structure simulations are expanded to regional simulations that include hundreds of thousands of structures [8 - 11]. To conduct such a task for a seismic event, the complete time histories of ground motions at all structural locations, as well as each structure's physical properties (at least mass, stiffness, and damping), are needed. In addition to scarcity, such detailed information is highly uncertain due to construction practices, material irregularities, and modeling limitations.

Due to the issues discussed above, other approaches are often adopted to assess the infrastructural damage from data and observations obtained directly after the hazard. The well-known structural health monitoring (SHM) techniques fall into this category. Advances in SHM demonstrate structural damage detection capability using measurements from sensors deployed in structures [12 – 15]. However, this option is not yet commonly available due to the high cost of sensor installation and maintenance. At the same time, on-ground field observations are conducted for a comprehensive assessment of post-earthquake infrastructural damage. Post-event reconnaissance surveying teams are commonly dispatched to the affected region to inspect damage and failure mechanisms of buildings [16]. Although valuable information is gathered in these surveys, the excessive time requirements and lack of guidelines for identifying the information-rich buildings for inspection make this method inefficient. Consequently, only a limited number of buildings, selected randomly, are chosen for a damage inspection.

## 1.2 Regional Building Damage Inference Problem

Buildings in a region share various aspects that are influential in their susceptibility to earthquake damage, and certain patterns can be found in the mechanisms that cause buildings damages in a region [17 – 19]. In other words, different buildings' damage intensities can be inferred by entering different inputs to a single function. This problem can be solved as a black-box model with the building variables and ground motion characteristics as inputs and the building's damage state as output.

To understand regional damage distribution from the scarce observation data, the isolated information needs to be translated in a consistent and systematic way. One can utilize the data obtained from field observations to train a supervised model and infer the damage for unobserved buildings in the region. To maintain affordability, minimal experimental data should be demanded.

To fit this model, direct outputs of the black box are needed for different inputs. Therefore, a surrogate model can be considered to emulate this black-box function. In the case of continuous damage labels, a regression algorithm should be considered as the surrogate model. As shown in [20, 21], Gaussian process regression (GPR) can be a suitable surrogate for modeling the non-linear development of buildings' damage when the observation data is sparse.

In this problem, observations are obtained as the reconnaissance team physically inspect buildings on a sequential basis. An active learning procedure can handle the efficient sampling of these inputs to fit the model with a minimal number of label queries. Various active learning kriging methods are proposed for structural reliability analysis where the areas located near the boundary of the failure domain is of high value for sampling [22-25]. In the regional damage estimation problem, however, the interest is in the prediction of damage for all buildings. Therefore, the damage labels of the buildings in the pool have a uniform level of importance, and the question is how to sample the buildings from the pool such that the prediction accuracy is maximal for the rest of the pool. Fig. 1 shows an active learning strategy for regional damage assessment of buildings.

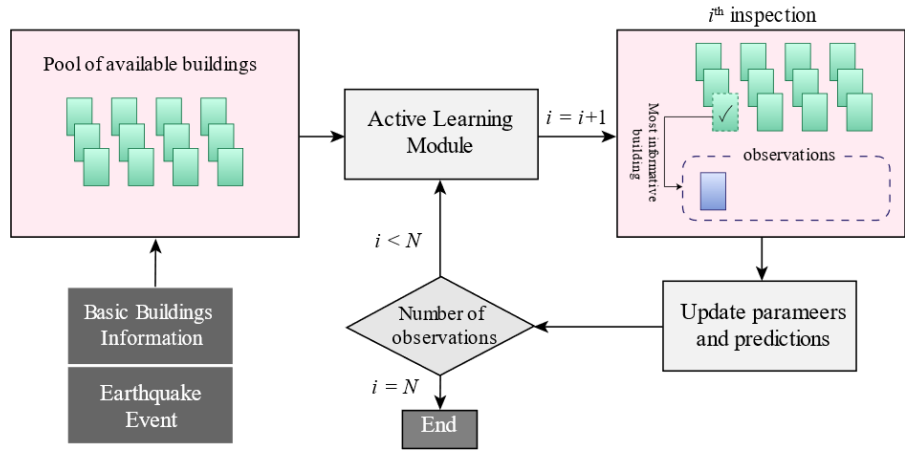

Fig. 1 The application of the active learning framework for the regional damage prediction problem

In a pool-based active sampling approach, a simple heuristic is to consider the prediction confidence interval as a measure to find the datapoints that the model is most uncertain about. This sampling method minimizes the model entropy [26]. A greedy algorithm was proposed in [27], referred to as the Active Learning McKay (ALM), where a sequential sampling is performed by picking the point with the highest variance as the next point for user labeling. However, this method tends to favor the datapoints close to the input domain boundary for sampling. In high-dimensional input spaces, boundary points mostly carry information about the outside of the domain of interest, resulting in a waste of information obtained from the queried labels [28]. Another popular algorithm was suggested by Cohn [29], where at each step, the expected reduction in predictive variance is maximized over the entire input space. Although showing better results compared to ALM, the Cohn method can be extremely expensive, computationally [30].

Guestrin et al. [28] proposed a sampling method based on the information-theoretic quantity of Mutual Information (MI) to address the issue of picking points on the edges of the input domain in ALM. MI measures the amount of information in terms of the Shannon entropy provided from observation of a random variable about another [31]. This idea can be applied as a sampling strategy to find a query subset that maximizes the

MI between observed and unobserved sample outputs. It was shown that MI performs superiorly compared with other classical experiment design criteria [28]. MI has been widely studied in sensor placement [32], robot path planning [33-35], design of experiments [36], and pool-based sample selection [37, 38]. However, solving this optimization problem can be NP-hard. To that end, a greedy algorithm was proposed that would reduce the computations to polynomial complexity [28]. Nevertheless, the computational complexity of the MI algorithm has always been the objective of various studies. Beck and Guillas [39] introduced a nugget parameter for better stability and used block matrix formulations to reduce the matrix inverse calculations to a single Cholesky decomposition in each iteration. However, the memory requirements are still a limitation for a large pool of available datapoints. Other studies have developed batch mode active learning methods based on the MI criterion [37, 40].

To enhance the practicality of maximizing MI, in addition to the greedy algorithm, Gusterin et al. [28] suggested a local kernels (LK) strategy, which reduces the size of kernel matrices and hence improves the matrix inverse computations [28]. In the LK strategy, due to the negligible impact, datapoints with covariances smaller than a threshold are filtered out of the covariance matrices at each step. The hyperparameters required for creating the covariance matrices were estimated from the batch of training data in [28]. However, in the damage assessment problem, where the data becomes available on a sequential basis, the training data and the hyperparameters are not available before starting the damage inspection surveys. Therefore, if the standard LK algorithm is used, the set of intial hyperparameters should be chosen arbitrarily, which will result in suboptimal prediction performance.

In this paper, an adaptive local kernels (ALK) method is proposed that will learn the hyperparameters through maximum likelihood estimation (MLE) as new labels become available. As a result, the influence of the irrelevant dimensions of the dataset in the sampling procedure is reduced, and the representativeness of the chosen datapoints improves. This algorithm can be utilized to determine judicious choices of sparse structural observations iteratively. Continuing on the previous work in the regional infrastructural damage assessment after earthquakes [21], the active learning and sampling are established based on the Gaussian process regression algorithm. The framework is tested on an earthquake testbed which is simulated using the rWHALE program provided by the SimCenter at NHERI [41]. The performance of the adaptive inspection using different active learning algorithms is evaluated in both accuracy and computational efficiency. The results are also compared with offline learning, where the batch selection approach is taken.

The remainder of this article is as follows: Section 2 provides the basic theories of the GPR, active learning and the standard MI algorithms. Section 3 introduces the proposed adaptive local kernels method and demonstrates its advantages on a simplified example. Section 4 describes the earthquake testbed and the dataset that is used to show the application of the proposed method in regional damage assessment. Section 5 thoroughly evaluates the outcomes of the damage estimation with the proposed method and compares them with the current active learning methods. And section 6 summarized the contributions of this study.

## 2. Active Learning with Gaussian Process Regression

Surrogate models are popular for representing high fidelity computer experiments or physical phenomena. The calibration procedure for these models consists of sampling an experiment at various input values, which can be very expensive. Choosing these values randomly might be inefficient since redundant or uninformative instances might be selected [42]. The purpose of machine learning with active sample selection, also known as active learning, is to find inputs with a configuration that maximizes the metamodel's learning rate under a fixed budget. In a pool-based scenario, where a pool of $n$ unlabeled samples is given, and the training budget is limited to $h$ samples, active learning iteratively picks a maximum of $h$ instances so that the trained model can predict the labels for the remaining samples with maximum accuracy [43].

In this section, the theory of GPR and maximum likelihood estimation of its hyperparameters are briefly discussed. Also, the theoretical basis of active sample selection methods for GPR is introduced. Starting with the ALM method, the simple greedy algorithm based on the probabilistic predictions of GPR is described. Next, the formulation of the standard MI algorithm and its improvements over the ALM method are explained. The efficient LK algorithm and the proposed ALK method are introduced in the subsequent discussion.

## 2.1. Gaussian Process Regression

To predict the damage intensity of a building, linear regression can be performed. However, linear regression does not provide any notion of uncertainty for the predictions made [28]. GPs are natural generalizations of the linear regression model that can provide a measure of confidence for the prediction of any new input. GPR (also known as kriging in geostatistics) has been proven to be a powerful method for highly non-linear functions [44-47]. A GP is a distribution over functions such that any finite number of points sampled at particular inputs from each function has a joint Gaussian distribution [48]. The posterior distribution of these functions is calculated by conditioning a prior on the output observations. The posterior distribution at an input can then be considered as the predicted output for that input location. From a simple weight-space view, we can analogize the GP prediction to a linear regression by considering every prediction as a weighted sum of the observed outputs. The weights are determined based on the similarity of the inputs of the observed outputs to the input of the to-be-predicted point [49]. The similarity measure is calculated in kernel space using a covariance function $K$, which is described later. Therefore, if we consider the set of all inputs as $\mathcal{V}$, given the observations $\mathbf{y}_{\mathcal{A}}$ for a subset $\mathcal{A} \subset \mathcal{V}$, the conditional probability distribution of label $y_x$ for a new input $x \in \mathcal{V}$ can be computed as $p(y_x|\mathbf{y}_{\mathcal{A}})$ such that

$$\mu_{x|\mathcal{A}} = \mu_x + \Sigma_{x\mathcal{A}}\Sigma_{\mathcal{A}\mathcal{A}}^{-1}(\mathbf{y}_{\mathcal{A}} - \mu_{\mathcal{A}}), \tag{1}$$

$$\sigma^2_{x|\mathcal{A}} = K(x,x) - \Sigma_{x\mathcal{A}}\Sigma_{\mathcal{A}\mathcal{A}}^{-1}\Sigma_{\mathcal{A}x} \tag{2}$$

where $\Sigma_{\mathcal{A}\mathcal{A}} = K(\mathcal{A},\mathcal{A}) + \sigma_n^2 \mathbf{I}$, $\sigma_n^2$ is independent Gaussian noise of observations, and $\mathbf{I}$ is the identity matrix.

Although the prior mean function is commonly considered zero, the covariance function plays an intrinsic role in modeling the data [49]. The Squared Exponential covariance function is widely used for practical applications due to its smoothness and differentiability. However, in this study, the Rational Quadratic (RQ) covariance function $K_{\text{RQ}}$ is used, which is a scaled summation of the SE functions [48]. In a previous study, the Automatic Relevance Determination type of the RQ function was shown to model the damage data with the best fit [21]. Considering two multidimensional input vectors $\mathbf{r}_p$ and $\mathbf{r}_q$, which specify the location of the datapoints $p$ and $q$ in the input space, the $K_{\text{RQ}}$ between these points can be computed as

$$K_{\text{RQ}}(p,q) = \sigma_f^2 \left(1 + (\mathbf{r}_p - \mathbf{r}_q)^T \frac{\mathbf{M}}{2\alpha} (\mathbf{r}_p - \mathbf{r}_q)\right)^{-\alpha} \tag{3}$$

where $\sigma_f^2$ is the signal variance, $\mathbf{M} = \text{diag}(\mathbf{l})^{-2}$ in which $\mathbf{l}$ is the vector containing the characteristic length scales, and $\alpha > 0$ determines the shape of the function. The parameters in the vector of length scales $\mathbf{l}$ and the signal variance $\sigma_f^2$ determine the level of variability of the covariance function. Every element of vector $\mathbf{l}$ can be varied to set the correlation of points in the associated dimension of the data. Further information regarding the properties of different covariance functions and their parameters can be found in [21, 48]. The vector of hyperparameters $\boldsymbol{\theta} = \{\boldsymbol{l}, \sigma_f^2, \alpha, \sigma_n^2\}$ is tuned during the training procedure with maximum likelihood estimation (MLE).

### 2.1.1 Maximum Likelihood Estimation (MLE)

The free parameters in vector $\boldsymbol{\theta}$ should be tuned so that the covariance function can model the dataset optimally. Because of the fortunate tractability of the integrals for Gaussian processes with Gaussian likelihoods, Bayesian inference can be incorporated to obtain a closed-form equation for the marginal likelihood with regards to the hyperparameters. In practice, it is easier to minimize the negative log marginal likelihood. The log marginal likelihood can be written as

$$\log p(\boldsymbol{y}_{\mathcal{A}}|\boldsymbol{r}_{\mathcal{A}}, \boldsymbol{\theta}) = -\frac{1}{2}\boldsymbol{y}_{\mathcal{A}}^T \boldsymbol{\Sigma}_{\mathcal{A}\mathcal{A}}^{-1} \boldsymbol{y}_{\mathcal{A}} - \frac{1}{2}\log|\boldsymbol{\Sigma}_{\mathcal{A}\mathcal{A}}| - \frac{n}{2}\log 2\pi \qquad (4)$$

The optimization for this loss function is performed by a conjugate gradient algorithm where the partial derivatives of Eq. 4 with respect to elements of $\boldsymbol{\theta}$ can be calculated as

$$\frac{\partial}{\partial \theta_j}\log p(\boldsymbol{y}_{\mathcal{A}}|\boldsymbol{r}_{\mathcal{A}}, \boldsymbol{\theta}) = \frac{1}{2}\boldsymbol{y}_{\mathcal{A}}^T \boldsymbol{\Sigma}_{\mathcal{A}\mathcal{A}}^{-1} \frac{\partial \boldsymbol{\Sigma}_{\mathcal{A}\mathcal{A}}}{\partial \theta_j} \boldsymbol{\Sigma}_{\mathcal{A}\mathcal{A}}^{-1} \boldsymbol{y}_{\mathcal{A}} - \frac{1}{2}\mathrm{tr}\left(\boldsymbol{\Sigma}_{\mathcal{A}\mathcal{A}}^{-1} \frac{\partial \boldsymbol{\Sigma}_{\mathcal{A}\mathcal{A}}}{\partial \theta_j}\right) \qquad (5)$$

For the RQ covariance function, the directional derivatives of the log marginal likelihood with regards to the hyperparameters can be obtained as:

$$\frac{\partial \boldsymbol{\Sigma}_{\mathcal{A}\mathcal{A}}}{\partial l_i} = \sigma_f^2 Q^{-\alpha-1} \times \frac{|r_{p,i}-r_{q,i}|^2}{l_i^3} \qquad (6)$$

$$\frac{\partial \boldsymbol{\Sigma}_{\mathcal{A}\mathcal{A}}}{\partial \sigma_f^2} = 2\sigma_f Q^{-\alpha} \qquad (7)$$

$$\frac{\partial \boldsymbol{\Sigma}_{\mathcal{A}\mathcal{A}}}{\partial \sigma_n^2} = 2\sigma_n \mathbf{I} \qquad (8)$$

$$\frac{\partial \boldsymbol{\Sigma}_{\mathcal{A}\mathcal{A}}}{\partial \alpha} = \sigma_f^2(-\ln Q + Q^{-1}) \times Q^{-\alpha} \qquad (9)$$

where $Q = 1 + (\boldsymbol{r}_p - \boldsymbol{r}_q)^T \frac{\mathbf{M}}{2\alpha}(\boldsymbol{r}_p - \boldsymbol{r}_q)$. Some libraries of the GPML toolbox [50] are used for inference purposes in this article.

## 2.2. Active Sample Selection Methods

In situations where labeling is expensive, and a pool of abundant unlabeled inputs is available, one should carefully choose the input samples. In the regional damage assessment problem, if there are $n = |\mathcal{V}|$ buildings in a region, we can only inspect $h = |\mathcal{A}| \ll n$ for damage evaluation due to the time and budget restrictions. Any building inspection can take up to two man-hours, as reported in [51]. Therefore, to achieve the most information gain in an affordable time frame, the learning procedure should be able to suggest the most informative cases for label queries. An active learning strategy can be pursued to ensure the informativeness of the buildings that will be inspected for damage.

### 2.2.1 Active Learning MacKay (ALM) Method

A heuristic approach is to pick the inputs that are most uncertain about each other, which is equivalent to maximizing the entropy of the chosen inputs [27].

$$\underset{\mathcal{A}:|\mathcal{A}|=h}{\operatorname{argmax}} H(\mathcal{A}) \tag{10}$$

Solving this optimization problem, however, is NP-hard. Given a set of observations $\mathcal{A}$, the differential entropy of a Gaussian random variable $x$ is defined as

$$H(x|\mathcal{A}) = \frac{1}{2}\log(2\pi e \sigma^2_{x|\mathcal{A}}) \tag{11}$$

which is a monotonic function of the variance and can be obtained from Eq. (2). Therefore, instead of solving Eq. (10) in one operation, a greedy algorithm can be adopted to make the observations sequentially. In this approach, at each step, given the indices of the points labeled previously, the unlabeled point with the highest variance is picked for label query [27].

The most uncertain areas in a fitted GP are the boundaries of the input domain. Therefore, picking the inputs where they are most uncertain about each other is also equivalent to placing the observations on the boundary of the input domain. Since every observation can improve the prediction uncertainty for queries within close proximity of its input, choosing the samples on the boundaries wastes a great portion of the observation information [28]. This issue is more pronounced in high dimensional datasets.

### 2.2.2 Standard Mutual Information (MI) Method

To adjust the problem of boundary preference in the ALM method, it was proposed to maximize the MI between observed and unobserved samples, rather than focusing on maximizing the entropy only for the selected inputs [28]. This approach finds a set of samples such that the entropy of the training set is maximized while the entropy of the testing set is minimized. Intuitively, this approach gathers the most diverse and information-rich samples for training and leaves the testing set with datapoints that display smooth variations. This set is obtained by solving the following equation

$$\mathcal{A} = \underset{\mathcal{A}:\mathcal{A}\subset\mathcal{V}}{\operatorname{argmax}} H(\mathcal{V}\backslash\mathcal{A}) - H(\mathcal{V}\backslash\mathcal{A}|\mathcal{A}), \tag{12}$$

which is equivalent to maximizing the MI, $I(\mathcal{A}; \mathcal{V}\backslash\mathcal{A})$, between set $\mathcal{A}$ and the rest of the samples $\mathcal{V}\backslash\mathcal{A}$. Analogous to Eq. (10), this problem is NP-complete. To reduce the complexity, an approximation approach was proposed in [28], where a greedy algorithm finds a suboptimal subset $\mathcal{A}$ in poly-time. At each step, this algorithm picks the samples that increase the MI the most. After a few simplifications, the following equation can be used to pick the sample $x$ at each step

$$\underset{x: x\in\mathcal{V}\backslash\mathcal{A}}{\operatorname{argmax}} H(x|\mathcal{A}) - H(x|\bar{\mathcal{A}}), \tag{13}$$

where $\mathcal{V}\backslash(\mathcal{A}\cup x)$ is denoted as $\bar{\mathcal{A}}$. The first term in Eq. (13) is similar to the greedy entropy rule in Eq. (11), but the second term biases the objective towards the center of the input space. This algorithm is guaranteed to perform within a constant factor approximation of the problem in Eq. (12). However, the bottleneck is the computation of $\Sigma^{-1}_{\bar{\mathcal{A}},\bar{\mathcal{A}}}$ in Eq. (2) for all the samples in $\mathcal{V}\backslash\mathcal{A}$. This brings the computing complexity to $\mathcal{O}(hn^4)$, which is still impractical for large datasets ($n \gtrsim 1000$).

### 2.2.3 Mutual Information with Local Kernel (MI-LK)

To further improve the computational efficiency, it was suggested that only local kernels be considered in the calculations. In positive covariance functions, correlations between points decay with distance, and points located far from each other are practically independent [28]. Therefore, we can assume $H(x|\mathcal{A}) \approx \tilde{H}_\varepsilon(x|\mathcal{B})$, where $\mathcal{B}$ is obtained from removing all elements $x'$ from $\mathcal{A}$ if $|K(x,x')| < \varepsilon$ for a small $\varepsilon$. It was shown that for $d = |\mathcal{B}|$, the computing complexity reduces to $\mathcal{O}(nd^3 + hn + hd^4)$. To limit the computing complexity to a certain level, the size of the local kernel matrix, $d$, can also be restricted. This implementation enforces a limit on the maximum number of local neighbors for each $x^*$ such that $|N(x^*; \varepsilon)| \leq d$. The local kernels algorithm suggested in [28], referred to as MI-LK hereafter, is shown in Algorithm 1.

One of the shortcomings of the formulation of the MI-LK algorithm is that the process should take place before the label observation starts. In this algorithm, the vector $\boldsymbol{\delta}$ contains the values of local MI based on the defined kernel parameters. Since the scale of $\boldsymbol{\delta}$ is retained in the sequential process of this algorithm, and only parts of it are updated in each step, a set of constant hyperparameters should be used throughout the process. Therefore, it is not feasible to update the kernel hyperparameters for calculating MI values as new labels are seen in the learning procedure.

---
**Algorithm 1** MI with local kernels (MI-LK) [28]
**Input:**
    Covariance $\boldsymbol{\Sigma}_{\mathcal{V}\mathcal{V}}, h, \mathcal{V}, \varepsilon > 0$
**Output:**
    The set of recommended cases $A \subseteq \mathcal{V}$
**Begin**
1    $\mathcal{A} = \phi$
2    **for** $x$ **in** $\mathcal{V}$:
3        $\delta_x = H(x) - \tilde{H}_\varepsilon(x|\mathcal{V}\backslash x)$
4    **for** $j$ **in** $\{1, 2, \ldots, h\}$:
5        $x^* = \operatorname*{argmax}_x \delta_x$
6        $\mathcal{A} = \mathcal{A} \cup x^*$
7        **for** $x$ **in** $N(x^*; \varepsilon)$:
8            $\delta_x = \tilde{H}_\varepsilon(x|\mathcal{A}) - \tilde{H}_\varepsilon(x|\bar{\mathcal{A}})$

---

## 3. MI with adaptive Local Kernels (MI-ALK)

In practice, the hyperparameters are rarely known before seeing any labels and are continuously updated by MLE as new labels are observed. It has been shown that by using the updated hyperparameters in each step of the standard MI procedure, the sampling performance improves significantly [38]. In this study, we modified the LK formulation to enable the possibility of the adaptable kernels to the hyperparameters' changes. The proposed adaptive local kernels method, referred to as MI-ALK, keeps the computing complexity within the practical ranges and does not require the known hyperparameters before seeing any instances. Algorithm 2 shows the steps of the procedure.

---
**Algorithm 2** MI with adaptive local kernels (MI-ALK)
**Input:**
    Covariance function $K, h, \mathcal{V}, \varepsilon \in [0,1)$
**Output:**
    The set of recommended cases $A \subseteq \mathcal{V}$
**Begin**
1    $\mathcal{A} = \phi$

| | |
|---|---|
| 2 | Initialize $\theta$ randomly |
| 3 | **for** $j$ in $\{1, 2, \dots, h\}$: |
| 4 | $\boldsymbol{\Sigma}_{\mathcal{V}\mathcal{V}} = \boldsymbol{K}_\theta(\mathcal{V}, \mathcal{V})$ |
| 5 | $\lambda = \boldsymbol{K}_\theta(x, x) \times \varepsilon$ |
| 6 | $x^* = \underset{x \in \mathcal{V}\backslash\mathcal{A}}{\operatorname{argmax}} H(x\|\mathcal{A}) - \widetilde{H}_\lambda(x\|\mathcal{V}\backslash(\mathcal{A} \cup x))$ |
| 7 | $\mathcal{A} = \mathcal{A} \cup x^*$ |
| 8* | $\boldsymbol{\theta} = \underset{\boldsymbol{\theta}}{\operatorname{argmax}} \log p(\boldsymbol{y}_\mathcal{A}\|\mathcal{A}, \boldsymbol{\theta})$ |
| 9* | label inference for $\mathcal{V}\backslash\mathcal{A}$ using Eqs. 1 and 2 |

\* These lines are a part of GPR algorithm in general

In contrast to the LK algorithm, the ALK does not rely on the initial hyperparameters, and an arbitrarily chosen initial $\boldsymbol{\theta}$ vector does impact the algorithm's performance. The algorithm starts with the allocation of a null set to $\mathcal{A}$ and iterates through lines 3 to 9 for a maximum number of $h$ iterations. In each iteration, the covariance matrix is updated with the new hyperparameter set obtained from MLE based on the seen labels. To keep the close neighbors of $x$ in the calculations of the local kernels, we can define a parameter $\lambda$ as a percentage of the maximum correlation by multiplying the coefficient $\varepsilon$ to the autocovariance of any point in the dataset. Line 6 is responsible for the determination of the most informative point given the set of training points $\mathcal{A}$. To avoid large sets of $N(x; \varepsilon)$, at this step, a limit can be imposed on the maximum number of neighbors. The label is queried for the chosen point, and $\boldsymbol{\theta}$ is updated through MLE at line 8. It should be pointed out that lines 8 and 9 are essential parts of any sequential GPR procedure and are not specific to this algorithm. We only use the results of the MLE at line 8 to improve the covariance calculations in our datapoint selection with MI. Depending on the approach for limiting the number of neighbors, the computing complexity of this algorithm is always $\leq \mathcal{O}(nhd^3)$.

### 3.1 Concept validation on a non-linear system response prediction

In order to validate the improvement of the active sample selection quality with the proposed adaptive kernels method, the LK and ALK algorithms are compared on predicting the maximum response of a simple non-linear SDOF system in this section.

**Numerical Model Formulation:** A Bouc-Wen (BW) model is used to generate a dataset with multiple input parameters determining the shape of the hysteresis cycles of the SDOF system. BW is a phenomenological model that is experimentally validated and shown to be able to capture the nonlinear behavior of inelastic steel material [52, 53]. The BW formulation is based on the displacement $u$ and the restoring force of the SDOF system $z$ as [54]

$$\dot{z} = \frac{1}{\eta}[A\dot{u} - v(\beta z|z|^{w-1}|\dot{u}| + \gamma|z|^w \dot{u}]  \tag{14}$$

where $\eta, A, v, \beta, \gamma,$ and $w$ determine the shape of the hysteretic behavior of the system. Rearranging the parameters, we have:

$$\dot{z} = s_1 \dot{u} - (s_2 z|z|^{s_4-1}|\dot{u}| + s_3|z|^{s_4}\dot{u})  \tag{15}$$

where

$$\boldsymbol{s} = [s_1 = \frac{A}{\eta}, s_2 = \frac{v\beta}{\eta}, s_3 = \frac{v\gamma}{\eta}, s_4 = w].  \tag{16}$$

It has been shown that the acceptable range of parameters for vector $\boldsymbol{s}$ should be

$$\{\boldsymbol{s} \in \mathbb{R}^4 | \ s_1 > 0, |s_3| \leq s_2, s_4 \geq 1\}. \tag{17}$$

Based on this model, the equation of motion for a non-linear damped SDOF system as shown in Fig. 2 can be formed as

$$m\ddot{u} + c\dot{u} + ku + z = F(t) \tag{18}$$

where $m$, $c$, and $k$ are system's mass, damping, and stiffness, respectively, and $F(t)$ is a function of time that corresponds to input loading.

To generate the dataset, system parameters $m$, $c$, and $k$ are considered to be the same among generated models, and only the parameters in vector $\boldsymbol{s}$ are randomly chosen to populate the dataset. The values considered for these parameters are shown in Table 1.

**Training set generation:** The input loading to the system is set as $F(t) = 2\cos t$ over the time interval $t \in [0, 10]$. The maximum displacement response of the system, $\max_t u(t)$ is considered as the output label for prediction. We generated 400 realizations based on the random parameters described in Table 1. At this point, the dataset entails the four parameters of the vector $\boldsymbol{s}$ as input and the maximum displacement as output for every realization. However, the input dimensions chosen so far are direct variables of the system, and all of them are influential in the output of the system. In practical applications, however, unimportant or even pure noise dimensions might be present in the dataset. Therefore, to introduce noise in our dataset, copies of the parameters $s_1$ and $s_2$ contaminated with a zero-mean Gaussian noise of variance 0.0025, referred to as $s_1'$ and $s_2'$, along with two independent Gaussian noise variables with $\mathcal{N}(0, 1)$, referred to as $N_1$ and $N_2$, are added to the dataset. Consequently, the total number of the input dimensions increases to eight, as shown in Table 1. Similarly, a Gaussian noise of variance 0.0025 is added to the labels.

Table 1 System parameters used for generating the SDOF dataset. The symbols in parentheses indicate the corresponding length scale in the GPR model

| $m$ | $c$ | $k$ | $s_1$ ($l_1$) | $s_2$ ($l_2$) | $s_3$ ($l_3$) | $s_4$ ($l_4$) | $s_1'$ ($l_5$) | $s_2'$ ($l_6$) | $N_1$ ($l_7$) | $N_2$ ($l_8$) |
|---|---|---|---|---|---|---|---|---|---|---|
| 1.0 | 0.2 | 1.0 | $\sim \mathcal{U}(0.5, 2.5)$ | $\sim \mathcal{N}(0, 1)$ | $\sim \mathcal{N}(0, 1)$ | $\sim \mathcal{U}(1, 2)$ | $s_1 +$ $\sim \mathcal{N}(0, 0.0025)$ | $s_2 +$ $\sim \mathcal{N}(0, 0.0025)$ | $\sim \mathcal{N}(0, 1)$ | $\sim \mathcal{N}(0, 1)$ |

**Testing scenarios:** The hyperparameter vector $\boldsymbol{\theta}$ is used in the calculation of the covariance matrices in the MI procedure, as shown in Eq. (3). Therefore, providing an optimally chosen $\boldsymbol{\theta}$ for covariance calculations can improve the performance of MI. However, in a sequential learning scheme, $\boldsymbol{\theta}$ is not known before seeing any labels. Therefore, to demonstrate the impact of $\boldsymbol{\theta}$ on the sampling performance, the algorithms are tested with two different sets of initial $\boldsymbol{\theta}$. In case 1, for all realizations, the vector is chosen as $\boldsymbol{\theta} = [1, \dots, 1]_{1 \times 11}$, called as arbitrary $\boldsymbol{\theta}$. In case 2, $\boldsymbol{\theta}$ is determined based on the optimized hyperparameters from MLE by considering the entire dataset in training. To identify the optimal hyperparameters, the MLE is run for 100 trials with random seeds, as shown in Fig. 3. The distributions of the converged values for the SDOF system are shown in Fig. 3 over 100 trials. The converged hyperparameter set, which yields the largest MLE, marked with blue dots in Fig. 3, is considered as the optimum value. The outcomes of the algorithms based on this $\boldsymbol{\theta}$ are called MLE optimized $\boldsymbol{\theta}$.

Observing the obtained lengthscales for the 8 input dimensions of the dataset, it can be seen that smaller values are assigned to the first four dimensions, $l_1$ to $l_4$, compared to $l_5$ and $l_6$. This is expected since the latter dimensions are contaminated with noise, and larger lengthscales indicate a lower impact on the inference. This effect is magnified for dimensions $l_7$ and $l_8$ which are pure Gaussian noise variables.

The lengthscales are not known before seeing any labels, and therefore, active learning algorithms need to start with arbitrarily chosen initial $\boldsymbol{\theta}$, which is often a vector of constant values. If the active learning algorithm is not able to update the hyperparameters as new labels are seen, it will assume the same level of importance for every dimension of the dataset throughout the learning procedure, which results in suboptimal performance.

**Learning configuration:** To be consistent with the regional damage assessment conditions where a pool of buildings is available, and the reconnaissance team needs to choose from that pool, a transductive learning approach is pursued in this paper. In this approach, samples are selected from a pool of datapoints for label queries, and the outputs of the remaining samples in the same pool are predicted. To have meaningful results, each algorithm is run 100 times on 80% of the dataset, called $\mathcal{V}_{80}$, which is chosen randomly with different seeds. To have a fair comparison between all algorithms, it is better to include the training set in the prediction evaluations in order to reduce the effect of different testing sets on the results [43]. Therefore, the true labels of the training points are considered as the predictions for those points, and accuracy is calculated for all the points in the chosen $\mathcal{V}_{80}$.

For all scenarios mentioned above, the active learning algorithm chooses the samples sequentially using $\varepsilon = 0.01\,K(.,.)$ and $d = 50$. The accuracy of the predictions is evaluated using the standardized mean square error (SMSE) as

$$\text{SMSE}(\boldsymbol{\mu}_*, \boldsymbol{y}_*) = \frac{1}{n_*}\frac{\left(\sum_{i=1}^{n_*}(\mu_{*i}-y_{*i})\right)^2}{\text{var}(\boldsymbol{y}_*)} \tag{19}$$

and also, the correlation coefficient (CC) between the predicted and true labels as

$$\text{CC}(\boldsymbol{\mu}_*, \boldsymbol{y}_*) = \frac{1}{n_*-1}\sum_{i=1}^{n_*}\left(\frac{\mu_{*i}-\mathbb{E}(\boldsymbol{\mu}_*)}{\text{SD}(\boldsymbol{\mu}_*)}\right)\left(\frac{y_{*i}-\mathbb{E}(\boldsymbol{y}_*)}{\text{SD}(\boldsymbol{y}_*)}\right) \tag{20}$$

where SD indicates the standard deviation of the random variable.

**Comparison and results:** The performance of the algorithms is compared based on the initial feeding $\boldsymbol{\theta}$ in Fig. 4. This figure shows the median improvement of the prediction accuracy after the 10th sample is observed. Also, the values of the area under the curve (AUC) are shown to quantify the performance measurements.

Considering Fig. 4 (a) and (c), it can be seen that the LK algorithm can perform significantly better if optimal hyperparameters are used through the learning procedure. However, in the case of arbitrarily chosen hyperparameters, the samples chosen by the algorithm delay the convergence of the GPR. At the same time, Fig. 4 (b) and (d) show that the MI-ALK performs almost independently of the initial hyperparameters and provide samples that allow GPR to converge earlier. Comparing the AUC values in Fig 4 (c) and (d), it can be inferred that while the MI-LK performance improves about 20% if optimal hyperparameters are used, the MI-ALK changes only about 5% as a result of the different initial hyperparameters. Furthermore, comparing the cases with optimal initial $\boldsymbol{\theta}$ between the algorithms, it is observed that MI-ALK presents lower AUC, which indicates better performance.

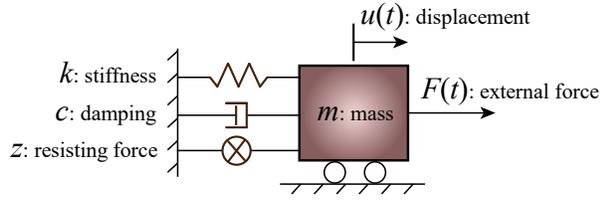
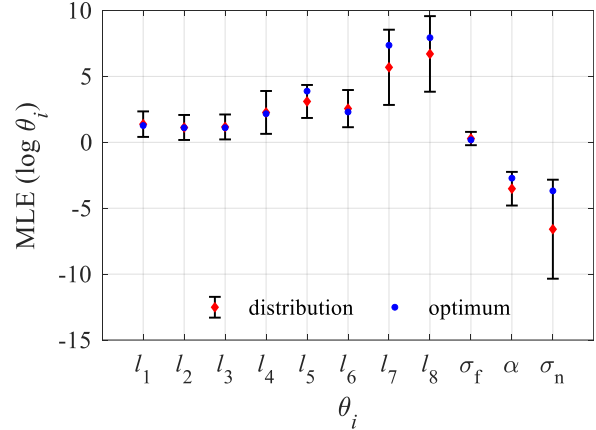

Fig. 2 The non-linear damped SDOF system setup

Fig. 3 Converged hyperparameters for the SDOF dataset using MLE

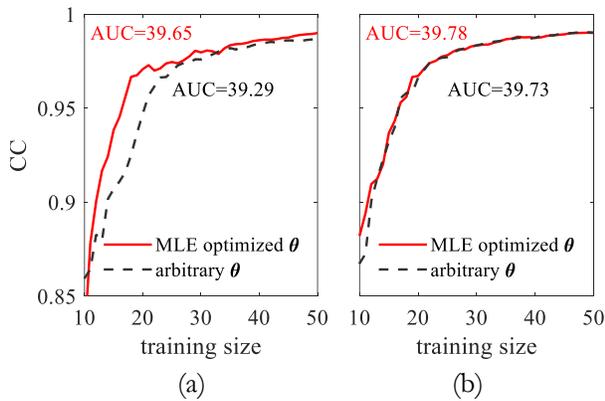
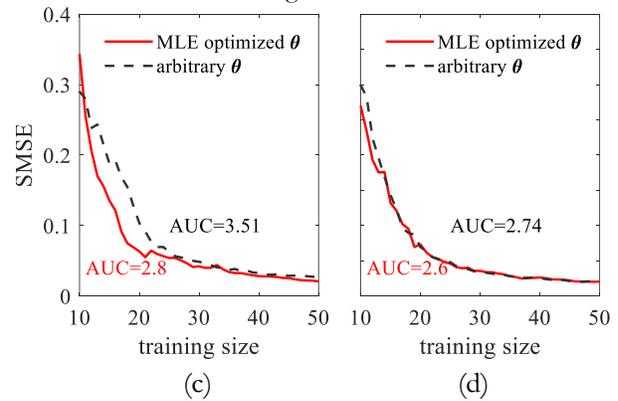

Fig. 4 The improvements in predictive performance for the SDOF dataset under different active learning scenarios. (a) and (c) show MI-LK results, while (b) and (d) present MI-ALK outcomes.

## 4. Regional Earthquake Impacted Building Damage Simulation Testbed

The application of the proposed data sampling method to a practical example is studied in this section. The case study consists of a simulated earthquake scenario using the rWHALE program provided by the SimCenter at NHERI [41]. The simulation is aimed at estimating the seismic damage and loss for individual buildings at a city scale. SimCenter designed a customizable workflow to streamline the risk assessment procedure. The workflow is briefly explained, and the earthquake testbed is described subsequently.

The workflow starts with gathering basic building information and creating a building inventory. The basic building information such as the number of stories, year built, floor area, structural type, etc., and other parameters such as the first vibration period suggested by [55] for typical building types are used to create an MDOF shear model for each building. The input ground motions to the MDOF models are chosen according to the geographical location of the buildings. A time history analysis is then performed to obtain the engineering demand parameters (EDP) such as maximum acceleration, drift ratio, and residual displacement. Once EDPs are calculated, a loss estimation procedure, adopted from the FEMA-P58 [56] guidelines, is followed and the economic loss ratio of the building along with repair time, repair cost, unsafe placard, etc. are calculated. More details regarding each step in the workflow are available at [41].

**Anchorage M7.1 earthquake:** Anchorage, AK, experienced a magnitude 7.1 earthquake on November 30th, 2018. This event is simulated in this scenario. Ground motions recorded by 38 strong-motion recording stations throughout Anchorage are obtained from [57] and are used as input to the models of 97k buildings (Fig. 5). A nearest neighbor algorithm is used to assign a ground motion to a building. To keep the visualizations optimal, 10k buildings are randomly chosen for this study.

### 4.1 Data description

Each of the 10k datapoints consists of basic building information and various earthquake intensity indices. In total, six variables are considered for each building, and six features are derived from ground motion signals. The labels are chosen from the typical EDPs of buildings to reflect different types of damage. While maximum absolute floor acceleration contributes to most non-structural damages in earthquakes, maximum drift ratio is an indicator of structural damage [58]. Also, the residual displacement is a proper measure for the repairability of the structure. Moreover, the economic loss ratio and the probability of unsafe placard of the building are computed as a result of the mentioned EDPs and are correlated to the total damage state of the building. A list of all features and labels are shown in Table 2. It should be noted that the occupancy type of buildings is a categorical variable and is one hot encoded in the input.

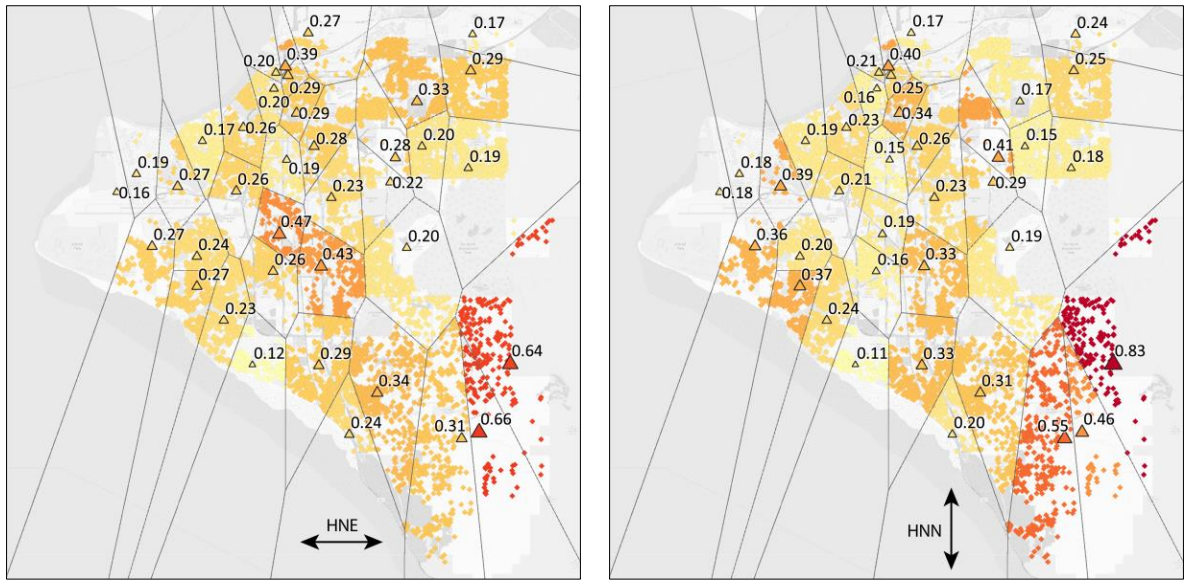

Fig. 5 Distribution of the seismic intensity in terms of PGA for NE and NN horizontal directions. Ground motion recording stations are shown with triangular markers where each station is assigned to its nearest buildings

Table 2 List of features and labels for the regional damage assessment dataset

| Building variables | | Earthquake indices | | Damage indices | |
| --- | --- | --- | --- | --- | --- |
| Feature | Range | Feature | Range | Label | Range |
| Floor area (m$^2$) | [25, 17723] | $S_a(T_1)$ (cm/s$^2$) | [93.3, 1323.9] | Max floor acceleration (m/s$^2$) | [1.6, 7.5] |
| Year of built | [1900, 2018] | Arias | [486.9, 4490.7] | Interstory drift ratio | [0.0, 0.02] |
| No. of stories | [1, 14] | Fajfar | [4.3, 20.7] | Residual roof displacement (m) | [0.0, 0.01] |
| Occupancy type | 5 types | Inter quantile range | [0.0, 0.2] | Unsafe placard probability | [0.0, 1.0] |
| Longitude | [-149.7, -150.0] | Kurtosis | [11.8, 110.1] | Economic loss ratio | [0.0, 0.9] |
| Latitude | [61.0, 61.2] | Spectral intensity | [357.1, 1156.3] | | |

### 4.2 Learning configurations

To assess the performance of the proposed algorithm in both computational complexity and prediction accuracy, a number of scenarios are considered for comparison. These scenarios are explained as follows:

**RND:** A batch learning approach with a random selection of the training set is assumed as a baseline for comparison with the active learning methods. This method is commonly used in general supervised learning applications.

**ALM:** As a simple greedy method, ALM is considered as the basic active learning approach in GPR. Although the computational complexity is very low, the predictive performance may not be great for high dimensional datasets.

**MI-LK:** The local kernels method (explained in Algorithm 1) helps the implementation of the MI criterion on large datasets by removing the less important datapoints from the covariance matrices. To keep the computing complexity at a certain range, for this problem, different values of $d$ and $\varepsilon$ are considered for the calculation of the local covariance matrices. The kernels are calculated with random selection for parameters of initial $\boldsymbol{\theta}$. Also, the independent noise $\sigma_n^2$ is considered to be $10^{-4}$ for all cases for better stability in the Cholesky decompositions.

**MI-ALK:** The adaptive local kernels method that is proposed in this study performs similarly to the MI-LK algorithm with the advantage of using label data to update the kernel space after each label query. In contrary to the MI-LK method, $\varepsilon$ is determined as a large percentage of the updated autocovariance. Similar to the MI-LK algorithm, both $\varepsilon$ and $d$ are altered for different computational demands and accuracy.

The summary of the considered algorithms, along with their parameter configurations, are shown in Table 3. For both MI-LK and MI-ALK algorithms, the two variants are chosen, such that the first represents an algorithm's fast performance while the accuracy might be compromised, and the second variant sets the parameters $\varepsilon$ and $d$ for larger and more flexible kernel matrices which results in greater computational demands and higher predictive performance.

Table 3 The studied learning methods and associated parameters

| Algorithm | Training steps | $\varepsilon$ | $d$ | No. of Random Testing Realizations |
|---|---|---|---|---|
| RND | 10:10:200 | - | - | 64 (random $\mathcal{V}_{80}$) |
| ALM | 1:200 | - | - | 64 (random initial $\boldsymbol{\theta}$ and $\mathcal{V}_{80}$) |
| MI-LK$_1$ | 1:200 | $10^{-2} \times K(.,.)$ | 100 | 64 (random initial $\boldsymbol{\theta}$ and $\mathcal{V}_{80}$) |
| MI-LK$_2$ | 1:200 | $10^{-5} \times K(.,.)$ | 800 | 64 (random initial $\boldsymbol{\theta}$ and $\mathcal{V}_{80}$) |
| MI-ALK$_1$ | 1:200 | 0.999 | 100 | 64 (random initial $\boldsymbol{\theta}$ and $\mathcal{V}_{80}$) |
| MI-ALK$_2$ | 1:200 | 0.95 | 300 | 64 (random initial $\boldsymbol{\theta}$ and $\mathcal{V}_{80}$) |

Since damage inspections during a short period after the occurrence of an earthquake cannot be performed in abundant numbers, the maximum training data size is considered to be 200 for this case study. The RND method is performed using increasing training set sizes at increments of 10. The active learning methods are assessed through a sequential approach where one label is queried at each step. In general machine learning applications, cross-validation of the training set is a suitable tool to provide insight into the performance of a model on the test set. However, cross-validation results may not be used as a fair measure to compare active learning methods. These methods tend to maximize the diversity of the selected samples in the training set and hence are expected to return poor cross-validation scores. In fact, it is highly likely that a random sampling approach scores higher cross-validation results compared to active learning methods, which is certainly not seen when evaluating the performance of the models on the testing set. Therefore, the evaluation of the algorithms' performances are done in the same way as the demonstrative example in section 3.1; a transductive learning on the $\mathcal{V}_{80}$ set. Also, to reduce the dependency of the results to the initial $\boldsymbol{\theta}$ and the pool of data, for every configuration, 64 realizations with random $\mathcal{V}_{80}$s and initial $\boldsymbol{\theta}$s are performed.

## 5. Inference Performance, Computational Efficiency and Discussion

The damage inference performance is evaluated based on several measures in this section. First, the levels of representativeness for samples chosen by each algorithm are presented. Second, the impact of the active learning algorithm on the accuracy of the predictions and the number of samples required for the GPR to infer with acceptable performance is evaluated. And third, the computational demands of the algorithms are compared.

### 5.1 Representativeness

A key factor in understanding the advantages of the sampled data in an active learning algorithm is the representativeness of the training data for the entire testing pool. Considering Eq. 1, it can be inferred that GPR predicts each new label by calculating a weighted average of the previously observed labels. The weights are calculated in the covariance matrix, and training points that are closer to the desired testing point have larger covariances. Therefore, one could expect higher predictive performance if testing points had closer representatives in the training set. To obtain a training set with high representativeness, an active learning method should select samples that can improve the prediction accuracy of the maximum number of unlabeled points located in close proximity to the selected samples [59].

To find the maximum similarity of a testing point to any point in the training set, we measure the covariance of each testing point to its closest point in the training set. The closer the measured correlation is to unity, the more similar a point is to its representative in the training set. The maximum similarity of the testing point $x^*$ at step $h$ of the training procedure can be obtained from the following equation

$$\text{Cov}_{\max}(x^*)_h = \max_{i \in \mathcal{A}(1:h)} K(x^*, x_i) \tag{21}$$

where $\mathcal{A}$ is the set of all training points. The final set of identified hyperparameters are used for covariance calculations in Eq. 21.

In this section, firstly, we use a random batch sampling approach to demonstrate the effect of representativeness on the prediction accuracy. To measure the prediction accuracy from a sensible point of view for individual datapoints, the Relative Difference (RD) formula is used as

$$\text{RD}(\mu_*, y_*) = \frac{|(\mu_* - y_*)|}{\max(\mu_*, y_*)} \tag{22}$$

Fig. 6 shows the comparison of prediction accuracy for testing points when GPR is trained on randomly selected training sets of size 200 over 64 realizations. The normalized $\text{Cov}_{\max}$ is used to differentiate the well-represented testing points from the under-represented ones. The results shown in Fig. 6 indicate that there is a positive correlation between the level of representativeness and prediction accuracy. Based on this observation, we can state that an under-represented testing point is more likely to be predicted inaccurately compared to a well-represented testing point. Therefore, the $\text{Cov}_{\max}$ criterion can be used to assign a level of confidence to the prediction of a testing point when true labels are unknown.

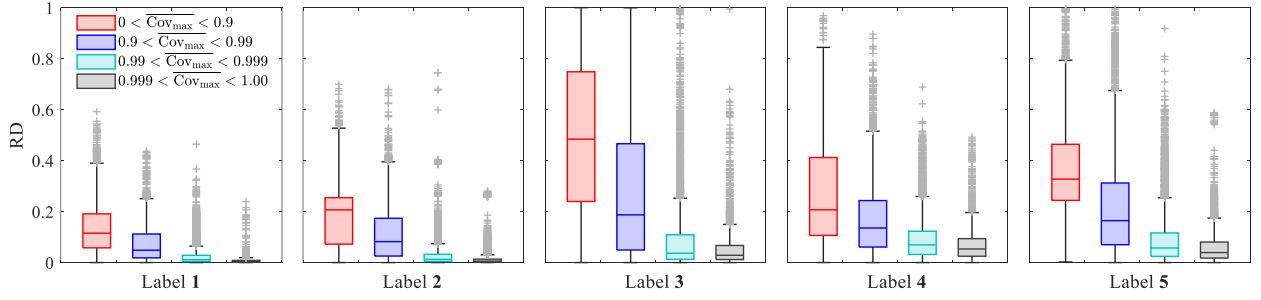

Fig. 6 Comparison of the prediction accuracy based on the normalized $\text{Cov}_{\max}$. Boxes show the interquartile ranges of the predictions based on the RD error.

Observing the effect of representativeness on prediction accuracy, we move on to compare the training sets sampled by active learning methods based on this criterion. Representativeness is calculated at each step of the active learning for all the points in the testing set to provide a better intuition into the behavior of each method. To obtain a smooth curve that shows the distribution of the $\text{Cov}_{\max}$ for all testing points, a normal distribution curve with $\mathcal{N}(\text{Cov}_{\max}(x^*)_h, 0.01)$ is considered for every testing point $x^*$, and summation of all the individual curves is considered as the distribution of the maximum training-testing similarity. Since the domain of similarity is limited to [0, 1], a truncated normal distribution should be considered where the probability density function (PDF) can be calculated as

$$\text{PDF}(a; \mu', \sigma, \rho_{\min}, \rho_{\max}) = \frac{1}{\sigma} \frac{\phi\left(\frac{a-\mu'}{\sigma}\right)}{\Phi\left(\frac{\rho_{\max}-\mu'}{\sigma}\right) - \Phi\left(\frac{\rho_{\min}-\mu'}{\sigma}\right)} \tag{23}$$

where $\mu'$ is the mean, $\sigma$ is the standard deviation (SD), $\rho_{\min}$ and $\rho_{\max}$ define the domain of the variable $\rho_{\min} \leq \alpha \leq \rho_{\max}$, and

$$\phi(\xi) = \frac{1}{\sqrt{2\pi}} \exp\left(-\frac{1}{2}\xi^2\right), \tag{24}$$

$$\Phi(\xi) = \frac{1}{2}\left(1 + \text{erf}\left(\frac{\xi}{\sqrt{2}}\right)\right). \tag{25}$$

Therefore, to obtain a smooth spike at the location of the maximum covariance for the testing point $x^*$, we calculate a PDF as

$$\text{PDF}(a; \text{Cov}_{\max}(x^*)_h, 0.01, 0, 1) = \frac{1}{0.01} \frac{\phi\left(\frac{a-\text{Cov}_{\max}(x^*)_h}{0.01}\right)}{\Phi\left(\frac{1-\text{Cov}_{\max}(x^*)_h}{0.01}\right) - \Phi\left(\frac{0-\text{Cov}_{\max}(x^*)_h}{0.01}\right)} \tag{26}$$

and to obtain the distribution of the maximum covariances for all testing points at step $h$, we have

$$\text{PDF}(\text{Cov}_{\max}(X^*)_h) = \frac{1}{n} \sum_{i=1}^{n} \text{PDF}(\text{Cov}_{\max}(x_i^*)_h) \tag{27}$$

where $n$ is the total number of testing points. Finally, this procedure is repeated for every realization and averaged over all realizations, so we can rewrite Eq. 27 as:

$$\text{PDF}(\text{Cov}_{\max}(X^*)_h) = \frac{1}{n'n} \sum_{j=1}^{n'} \sum_{i=1}^{n} \text{PDF}(\text{Cov}_{\max}(x_i^*)_h)_j \tag{28}$$

where $n'$ is the number of realizations. Eq. 28 is used for each active learning configuration in Table 3 and for all labels.

The calculated PDFs at step $h = 50$ of each algorithm are shown in Fig. 7 to characteristically compare the obtained similarities of each algorithm at the early steps of the training procedure. The differences in the density of the curves on small similarity ranges are noticeable between MI-LK and MI-ALK variants. MI-LK variants present higher densities in the similarity range of [0, 0.5] for all five labels, which indicates that a higher percentage of testing points are located within this range. We consider these points as poorly-represented in the training set. To quantitively compare the algorithms, the percentage of the poorly-represented points at step $h = 50$ of the training procedure are shown in Table 4. Comparing MI-LK with MI-ALK variants, it can be seen that the percentage of poorly-represented testing points is between 0.08% and 2.68% for MI-ALK, while this percentage is between 0.51% and 7.87% for MI-LK. It should be noted that although ALM seemingly presents good results from this point of view, on average, about 70% of the testing points are located below 0.9 similarity, and thus, ALM still performs poorly considering the percentage of the very well-represented testing points. This percentage is about 20% for MI-LK$_1$ and MI-ALK$_1$ and about 16% for MI-LK$_2$ and MI-ALK$_2$.

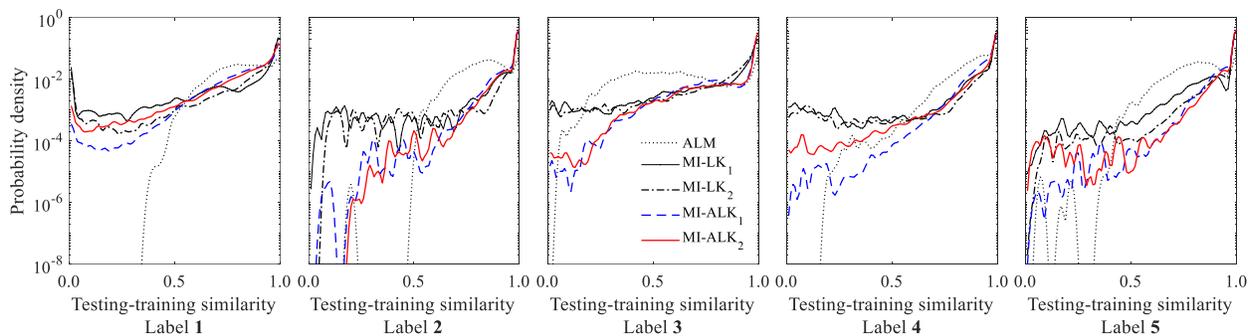

Fig. 7 The PDFs of maximum similarity at step $h = 50$

Table 4. The average percentage of testing points with maximum similarity less than 0.5 among training points at step $h = 50$

| Algorithm | ALM | MI-LK$_1$ | MI-LK$_2$ | MI-ALK$_1$ | MI-ALK$_2$ |
|---|---|---|---|---|---|
| Label 1 | 0.17 | **7.87** | 4.31 | 0.95 | **2.68** |
| Label 2 | 0.00 | 2.70 | 2.39 | 0.11 | 0.10 |
| Label 3 | 35.87 | 6.79 | 5.92 | 2.08 | 2.30 |
| Label 4 | 0.24 | 3.47 | 2.92 | 0.08 | 0.53 |
| Label 5 | 0.74 | 1.16 | **0.51** | **0.08** | 0.23 |

To visualize the variations of the PDFs based on the arrival of the new labels, the obtained PDFs for each step of the training procedure are stacked side by side to form a 2D image and present the variations of the distribution of the maximum similarity for every configuration. Fig. 8 shows the progress of the maximum similarity during the training procedure. In this figure, the shaded areas show the PDFs from the top, where a darker color indicates a smaller probability density. We can see that for all configurations, the mean of the PDFs, which is shown by a dashed line, increases as new points are added to the training set. These modifications are stemmed from the increase in the number of testing points having close representatives, which improves the training-testing similarity.

Considering the shaded areas, we notice light-colored horizontal stripes on the images for the MI-LK methods that are stretched through the training procedure. These strips indicate a high density of low-represented testing points. Although the strips are present in MI-ALK configurations in label 2 and 3 as well, they fade away at training sizes smaller than 50, while they continue until above 150 training points in MI-LK configurations.

Furthermore, considering Label 1, narrow white bands at the bottom of the images for MI-LK$_1$ and MI-LK$_2$ are observable and highlighted with red rectangles. For instance, on average, after labeling 100 training points, about 3% of the testing points with MI-LK$_1$ and 1% of testing points with MI-LK$_2$ still have almost zero similarity to any points in the training set. We can also notice the slower rate of progress in similarity for the ALM method. For instance, using ALM for the prediction of label 3, Fig. 8 shows that PDFs' mean improves at a slow rate and even after observing 150 labels, about 37% of testing points have a maximum similarity below 80%.

For easier quantitative comparison, the mean and SDs of the PDFs are shown at $h = 50, h = 100,$ and $h = 150$ of the training procedure in Fig. 9 and Tables A1 and A2. Compared to ALM, the higher PDFs' means and thus better representatives in the training set are observable with all MI methods. Also, assessing the three training steps, it can be seen that compared to MI-LK, the mean values are higher, and the SDs are smaller in MI-ALK variants. For instance, considering the 100$^{th}$ step of training for Label 3, the mean and SD of maximum similarity are at 0.93 and 0.15 for MI-LK$_1$, respectively, whereas they are at 0.97 and 0.07 for MI-ALK$_1$. It can be inferred that MI-ALK$_1$ has found similar representatives for a higher percentage of testing points. It is worth noting that from the representativeness point of view, the improvements are less significant when parameters are tweaked within each algorithm.

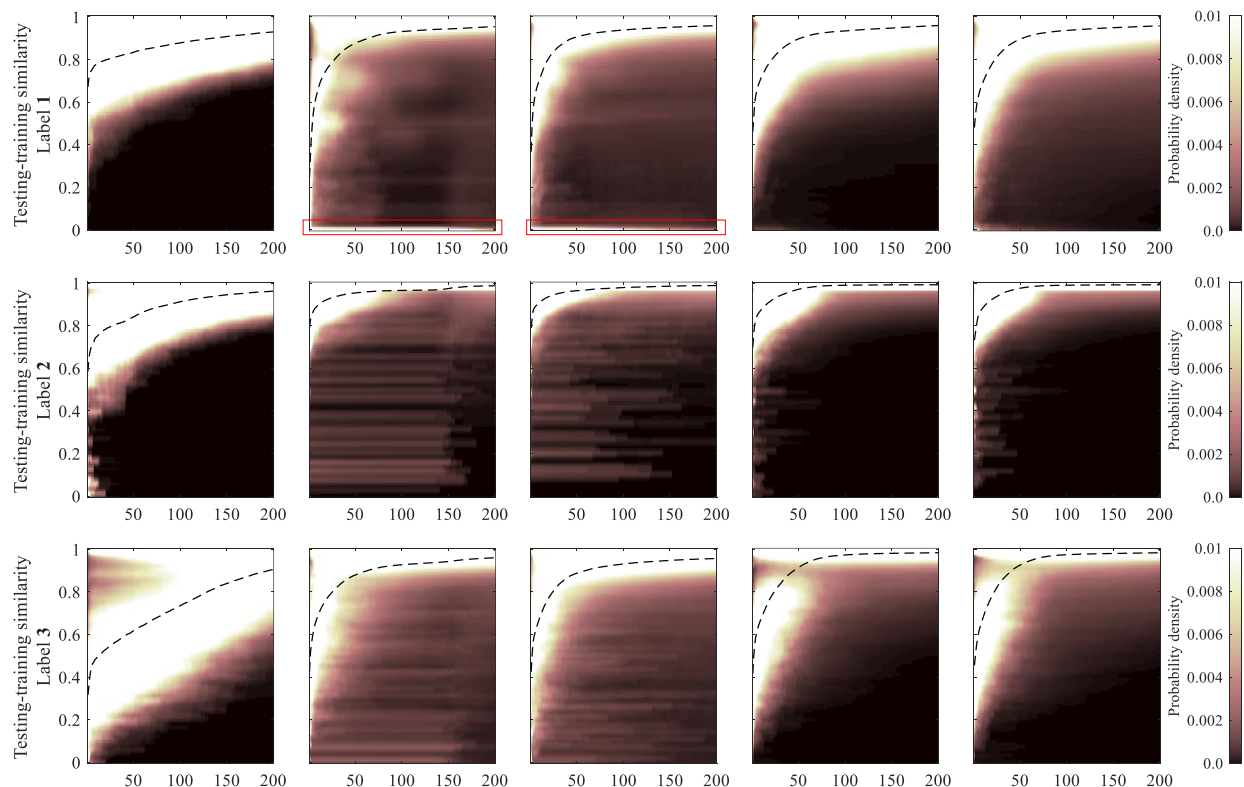

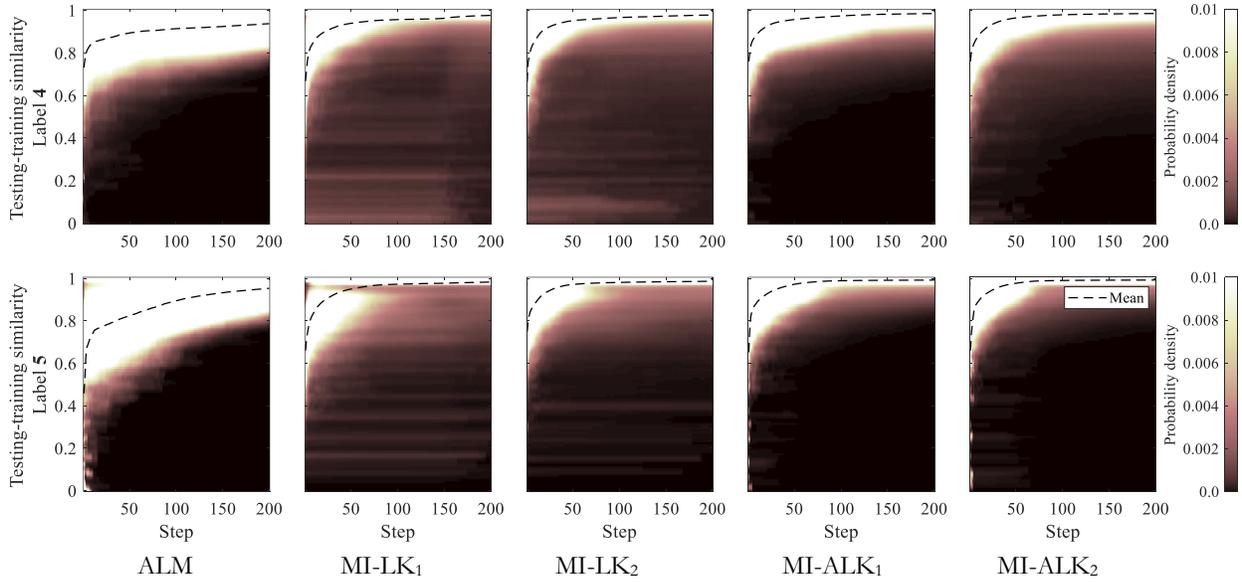
Fig. 8 The rate of improvement in the representativeness of the training points for each method

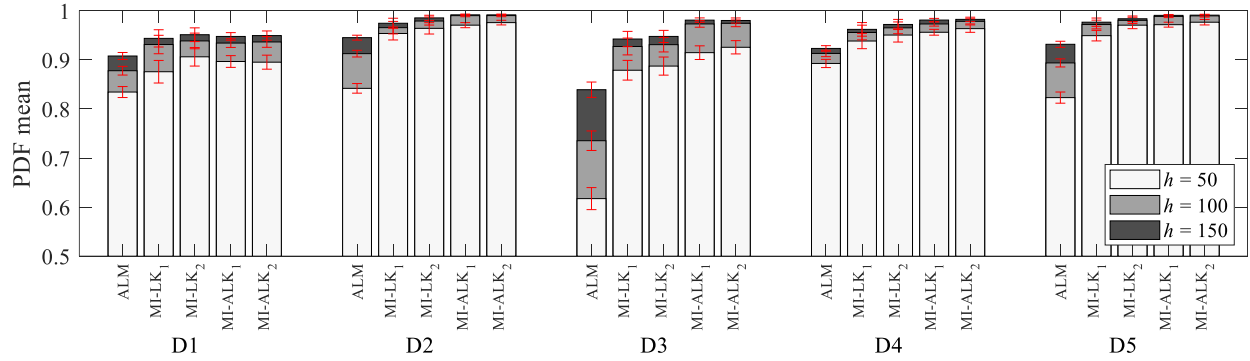
Fig. 9 Illustration of the mean and SD of the similarity PDFs at three different training steps

### 5.2 Predictive Performance

To compare the predictive performance across learning methods, SMSE and CC are calculated between the predicted labels and true labels. Figs. 10 and 11 show the calculated median of the progress of the predictive performance over all realizations as labels are queried for new samples in terms of the SMSE and CC, respectively. In these figures, methods that converge to an accuracy level with a smaller number of training points are preferred. It can be seen that as expected, the ALM method performs poorly compared to the other methods due to the issues stated in Section 2.2. The results of the random selection method (RND) are shown as error bars for each batch size. The error bars indicate the 0.25 and 0.75 quantiles, and it can be seen that high variations exist in the predictive performance of Label 3. These variations indicate the highly non-linear underlying function and vulnerability of GPR to the chosen training data for this label. At the same time, variants of the MI-LK method perform considerably better than the RND method for labels 1, 3, and 5. The MI-LK$_1$ shows less appealing results in the prediction of Label 2 and 5. While prediction results are improved with increasing the limit $d$ and reducing $\varepsilon$ in MI-LK$_2$ for all labels, both variants of MI-LK experience unstable predictions for Label 3.

On the other hand, compared to MI-LK, the two variants of MI-ALK present significantly better results for labels 2, 3, and 5, and converge after observing ~75 training points. The performance is comparable in Label 1 and 4 between MI-ALK$_2$ and MI-LK$_2$, although the latter requires a much higher limit for the covariance matrix size $d$.

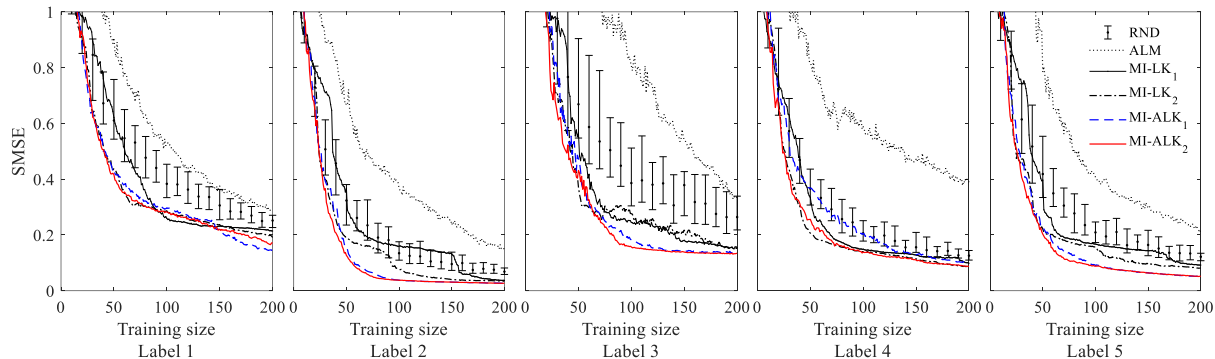

Fig. 10 Progress in the predictive performance of the GPR in terms of SMSE when using different active learning configurations.

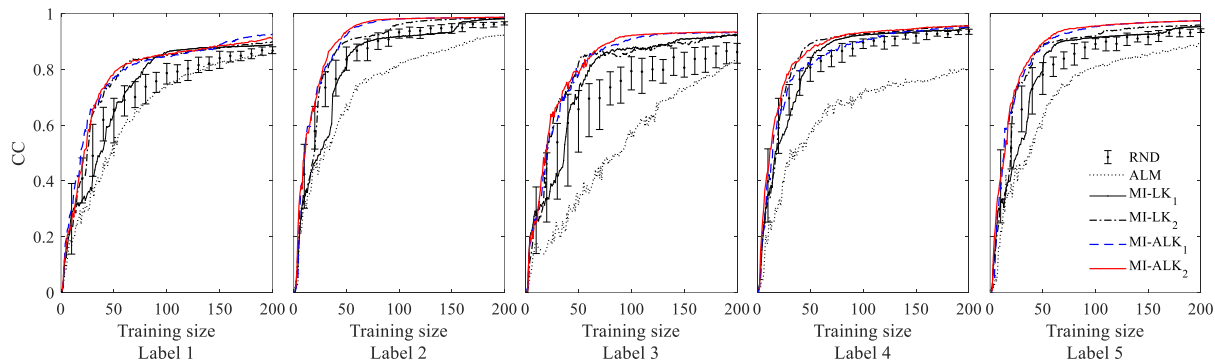

Fig. 11 Progress in the predictive performance of the GPR in terms of CC when using different active learning configurations.

### 5.3 Computational Complexity

At almost the same level of importance as the predictive performance, computing complexity should be evaluated for the proposed method. Computing complexity can be referred to as the total time demand of an algorithm as well as the memory requirements. In this study, the memory concerns are relieved by limiting the maximum number of local neighbors for both MI algorithms. However, one should be careful if using the standard greedy MI algorithm suggested in [28], as the inversion of large matrices are required. The focus of the computing complexity measurement in this study is the total processing time required by each method. To this end, a finer set of $\varepsilon$ and $d$ values are considered for MI-LK and MI-ALK algorithms to present the improvements in accuracy versus the time complexity. The additional sets of $\varepsilon$ and $d$ are shown in Table 5. Realizations are performed with parallel programming in Matlab and are run on Intel Xeon Skylake nodes with 32 cores each, allocated by the center for high performance computing (CHPC) of the University of Utah. The timings include the total time required by the sample selection procedure as well as the prediction steps until the maximum number of training points are selected.

Fig. 12 compares the processing times opposed by the predictive performance for each method. To quantify the predictive performance, the area under the curve (AUC) of the mean SMSE graphs are calculated for each realization [43]. The calculation of the AUCs starts at step 75, which is roughly the step that algorithms converge and present meaningful results. Also, for each method, the median value is shown with a larger marker in Fig. 12.

Table 5 Additional sets of $\varepsilon$ and $d$ for computational comlexty comparison

| Set number | | 1 | 2 | 3 | 4 | 5 | 6 | 7 |
|---|---|---|---|---|---|---|---|---|
| **MI-LK** | $\varepsilon \times K_{\boldsymbol{\theta}_0}(.,.)$ | $10^{-1}$ | $10^{-2}$ | $10^{-3}$ | $10^{-4}$ | $10^{-5}$ | $10^{-6}$ | $10^{-7}$ |
| | $d$ | 100 | 200 | 300 | 400 | 500 | 600 | 700 |
| **MI-ALK** | $\varepsilon$ | 0.9999 | 0.999 | 0.99 | 0.98 | 0.97 | 0.96 | 0.95 |
| | $d$ | 50 | 100 | 100 | 150 | 200 | 250 | 300 |

Comparing the results shown in Fig. 12, it can be seen that in all label predictions, the maximum accuracy obtained from MI-ALK is equal or higher than MI-LK. At the same time, this high accuracy is obtained with a better time complexity compared to the MI-LK's best performance. Comparing the median AUC SMSE for case 7 of both algorithms, the maximum performance of the MI-ALK algorithm is 4%, 47%, 32%, 3%, and 37% higher than MI-LK for labels 1 to 5, respectively. At the same time, the computational complexity of the MI-ALK for these labels is 25%, 7%, 16%, 14%, and 11% lower than MI-LK, respectively. Finally, at almost the same levels of computing complexity, it can be seen that case 1 of MI-LK performs significantly better than the ALM method.

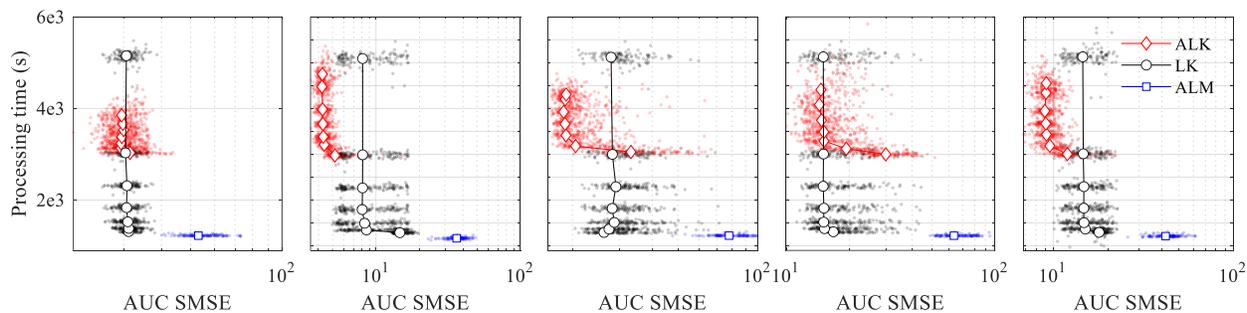

Fig. 12 Processing times required for each learning method to reach to the maximum training size limit. The x-axis displays the predictive performance in terms of AUC SMSE

### 5.4 Discussion

The proposed active learning method enables the GPR algorithm to be implemented in the damage assessment of buildings after an earthquake. This algorithm reduces the inspection costs by avoiding unimportant buildings and performs efficiently by finding a close representative for every building in the region. Therefore, in a progressive approach, the damage level of all buildings can be inferred with high accuracy.

Although the proposed algorithm provides more informative samples for the GPR algorithm and expedites the improvements of performance, it requires higher computational demands compared to the LK algorithm **for identical $\varepsilon$ and $d$ parameters**. This extra computation is well justified when label query comes at high expenses, such as in the regional damage assessment problem. In other words, to achieve the same level of predictive performance, we may identify the samples in a short time using the MI-LK algorithm and spend a certain amount of time querying the labels or identify fewer but more informative samples using the MI-ALK algorithm and consequently spend less time in the label query process. The fewer number of label queries can reduce the overall learning expenses and compensate for the extra computational demand.

Considering Figs. 10 and 11, although the gap between the results of LK and ALK methods diminishes after 200 training points for labels 2, 3, and 4, the final accuracy of the ALK is still higher in labels 1 and 5. In this article, we assumed a budget for 200 training points. However, if the limit were anywhere between 50 to 150 training points, the final predictive performance of the ALK method would be significantly higher than LK in labels 2, 3, and 5. The urgent demand for the reconnaissance data and the limitation of resources in the aftermath of an earthquake can cause such limitations, and therefore, a method that is known to provide higher performance using fewer damage inspections should be adopted. The ALK method can train a more powerful surrogate model under those limitations, which subsequently leads to a more accurate estimation of the overall losses due to an earthquake.

Finally, it is worth noting that the processing times shown in Fig. 12 are obtained when 32 realizations were run simultaneously in parallel. In a real-world scenario, a single realization on a laptop or PC with higher CPU clock speeds compared to the CHPC nodes (2.1 GHz) will require significantly shorter processing times. In fact, the calculations required for sample selection with MI-ALK$_2$ on an Intel Core i5 7500 CPU, only adds about ~4 seconds of computational overhead at each step. Therefore, without the need for strong workstations, the method can be applied on the go following the occurrence of an earthquake.

**6. Conclusion**

A new active-learning procedure is formulated to adaptively select and infer the post-seismic building damage in an impacted region. Through a non-linear SDOF response prediction test, it was concluded that updating the hyperparameters used to create the kernel matrices utilized by the MI formulation after observing new labels can adjust the sample selection by reducing the effect of the unimportant data dimensions. The key contributions and findings of the manuscript are described in the following:

- The adaptable formulation of the local kernels strategy based on the information theoretic measure of mutual information can substantially improve the sample selection phase of the learning procedure.
- Through a simulated earthquake testbed, the performance of the proposed MI-ALK method was compared with the standard MI-LK method for the sample selection of 5 different damage indicators. It was shown that the predictions obtained by MI-ALK converge to acceptable levels of accuracy using fewer training points. Furthermore, the instability of predictions with the observation of new labels was reduced with MI-ALK.
- Compared to MI-LK, the samples chosen by the MI-ALK can cover the domain of input faster and are better representatives of the pool of unlabeled samples.
- The improvements in performance were observed while the MI-ALK performed at lower computational demands. For the labels considered in the regional damage assessment study, the performance of the MI-ALK showed improvements of up to 47% while reducing the computational demands up to 25%.

**Appendix**

Table A1 Mean values of the maximum similarity distribution at different training steps for each algorithm

| Algorithm | Training step | Label 1 | Label 2 | Label 3 | Label 4 | Label 5 |
|---|---|---|---|---|---|---|
| ALM | 50 | 0.83 | 0.84 | 0.62 | 0.89 | 0.82 |
|  | 100 | 0.88 | 0.91 | 0.74 | 0.91 | 0.89 |
|  | 150 | 0.91 | 0.95 | 0.84 | 0.92 | 0.93 |
| MI-LK$_1$ | 50 | 0.88 | 0.95 | 0.88 | 0.94 | 0.95 |
|  | 100 | 0.93 | 0.97 | 0.93 | 0.96 | 0.97 |
|  | 150 | 0.94 | 0.97 | 0.94 | 0.96 | 0.98 |

| Algorithm | Training step | Label 1 | Label 2 | Label 3 | Label 4 | Label 5 |
|---|---|---|---|---|---|---|
| MI-LK$_2$ | 50 | 0.91 | 0.96 | 0.89 | 0.95 | 0.97 |
|  | 100 | 0.94 | 0.98 | 0.93 | 0.97 | 0.98 |
|  | 150 | 0.95 | 0.99 | 0.95 | 0.97 | 0.98 |
| MI-ALK$_1$ | 50 | 0.90 | 0.97 | 0.91 | 0.96 | 0.97 |
|  | 100 | 0.93 | 0.99 | 0.97 | 0.97 | 0.99 |
|  | 150 | 0.95 | 0.99 | 0.98 | 0.98 | 0.99 |
| MI-ALK$_2$ | 50 | 0.90 | 0.98 | 0.93 | 0.96 | 0.98 |
|  | 100 | 0.94 | 0.99 | 0.97 | 0.98 | 0.99 |
|  | 150 | 0.95 | 0.99 | 0.98 | 0.98 | 0.99 |

Table A2 SDs of the maximum similarity distribution at different training steps for each algorithm

| Algorithm | Training step | Label 1 | Label 2 | Label 3 | Label 4 | Label 5 |
|---|---|---|---|---|---|---|
| ALM | 50 | 0.11 | 0.10 | 0.22 | 0.08 | 0.11 |
|  | 100 | 0.09 | 0.07 | 0.20 | 0.07 | 0.08 |
|  | 150 | 0.07 | 0.05 | 0.16 | 0.06 | 0.06 |
| MI-LK$_1$ | 50 | 0.23 | 0.13 | 0.20 | 0.16 | 0.11 |
|  | 100 | 0.19 | 0.13 | 0.17 | 0.15 | 0.09 |
|  | 150 | 0.18 | 0.10 | 0.16 | 0.14 | 0.08 |
| MI-LK$_2$ | 50 | 0.19 | 0.12 | 0.18 | 0.14 | 0.07 |
|  | 100 | 0.16 | 0.08 | 0.15 | 0.12 | 0.06 |
|  | 150 | 0.14 | 0.05 | 0.12 | 0.10 | 0.06 |
| MI-ALK$_1$ | 50 | 0.12 | 0.05 | 0.14 | 0.06 | 0.05 |
|  | 100 | 0.09 | 0.02 | 0.07 | 0.04 | 0.02 |
|  | 150 | 0.08 | 0.02 | 0.05 | 0.03 | 0.02 |
| MI-ALK$_2$ | 50 | 0.14 | 0.05 | 0.14 | 0.08 | 0.06 |
|  | 100 | 0.11 | 0.02 | 0.07 | 0.05 | 0.02 |
|  | 150 | 0.09 | 0.02 | 0.05 | 0.04 | 0.02 |

## Acknowledgement

This material is based upon work supported by the University of Utah, and the National Science Foundation under award numbers 1839833 and 2004658. Any opinions, findings, and conclusions or recommendations expressed in this material are those of the author(s) and do not necessarily reflect the views of the National Science Foundation.